\documentclass[a4page,10pt]{article}
\usepackage[top=0.8in, bottom=0.8in, left=0.5in, right=0.5in]{geometry}
\usepackage{fancyhdr}
\renewcommand{\thispagestyle}[1]{}
\usepackage[pdftex]{graphicx}
\usepackage[utf8]{inputenc}
\usepackage{hyperref}
\usepackage{tabularx}
\usepackage{floatrow}
\newfloatcommand{capbtabbox}{table}[][\FBwidth]

\title{ODE - Augmented Training Improves \\
Anomaly Detection in Sensor Data from Machines}

\begin{document}
\author{Mohit Yadav$^1$, Pankaj Malhotra$^1$, Lovekesh Vig$^1$, K Sriram$^2$, and Gautam Shroff$^1$
%
%
\vspace{.3cm}\\
%
1- TCS Research, New-Delhi, India\\
2- IIIT, New-Delhi, India \\
%
}
\maketitle 

\begin{abstract}
\label{Abs}
Machines of all kinds from vehicles to industrial equipment are increasingly instrumented with hundreds of sensors. Using such data to detect ``anomalous" behaviour is critical for safety and efficient maintenance. However, anomalies occur rarely and with great variety in such systems, so there is often insufficient anomalous data to build reliable detectors. A standard approach to mitigate this problem is to use one class methods relying only on data from normal behaviour. Unfortunately, even these  approaches are more likely to fail in the scenario of a dynamical system with manual control input(s). Normal behaviour in response to novel control input(s) might look very different to the learned detector which may be incorrectly detected as anomalous. In this paper, we address this issue by modelling time-series via Ordinary Differential Equations (ODE) and utilising such an ODE model to simulate the behaviour of dynamical systems under varying control inputs. The available data is then augmented with data generated from the ODE, and the anomaly detector is retrained on this augmented dataset. Experiments demonstrate that ODE-augmented training data allows better coverage of possible control input(s) and results in learning more accurate distinctions between normal and anomalous behaviour in time-series. 
\end{abstract}

\section{Introduction}
\label{Intro}
Modern machines are increasingly instrumented with hundreds and even thousands of sensors.
Moreover, with the advent of what is being called the ``industrial internet'', such sensors are also able to regularly transmit their data to the component (e.g., engine) manufacturer, OEM (e.g., aircraft / car manufacturer) or even operator (e.g., airline / trucking company). Detecting ``anomalous" behaviour from such sensor data is important to be able to, a) indicate signs of degraded performance to trigger early maintenance, b) predict failures before they actually happen, i.e., \textit{prognostics}  and c) serve as indicators to forecast potential product recalls when aggregated at a population level. A number of algorithms for anomaly detection have been proposed using a variety of data arising from financial markets, diagnostic systems, biological systems, and various other sources \cite{gupta2014outlier}. Since normal data is more easily available, a typical anomaly detection algorithm learns the behaviour of ``normal'' data and uses deviations from it to determine anomalies \cite{review_nd,anomalyDetection_multi_dim_ts,LSTM_AD,LSTM_AD2}.

Traditionally, statistical techniques such as cumulative sum (CUMSUM) and exponentially weighted moving average (EWMA) over a time window have been applied to detect variation from the underlying distribution of normal data \cite{stat_AD}. Yet another set of approaches are based on one-class SVM learning \cite{svm_AD,enahnce_svm_AD}. Predictive approaches attempt to predict the time-series and detect anomalies by learning a threshold on prediction errors \cite{pred1_AD,pred2_AD}. An important disadvantage of most of these approaches is their dependence on the choice of window length. Recently, a predictive deep Long-Short-Term-Memory based Anomaly Detection (LSTM-AD) has been proposed \cite{LSTM_AD, LSTM_AD2} to overcome the requirement of a pre-specified context window or data pre-processing. LSTM-AD is a robust times-series prediction model, as it can capture: i) long term temporal correlations, ii) correlations across dimensions in a multivariate time-series, and iii) temporal patterns across different resolutions. In this paper, we utilize the LSTM-AD approach and provide its description in Section \ref{LSTM_AD}. 

For dynamical systems, ODE models can leverage domain information and also take advantage of data-driven machine-learning methodologies \cite{LFM}. However to the best of our knowledge, hybrid strategies that marry ODE with machine learning have not made use of the inherent generative capability of ODEs. In this paper, we propose that an ODE based model can be used as a generative model for time-series data to overcome the difficulty in learning and predicting time-series when only a limited amount of normal data is available. Small volumes of normal data often contain insufficient variations of manual control input(s), thereby limiting the behaviour patterns that can be learned. An ODE based model can be used to augment data by simulating novel manual control input(s) not present in the available data. We show that when such augmented data is fed to LSTM-AD it learns a better discriminator between normal and anomalous behaviour. We assume that the structure of the ODE model is available from domain knowledge and estimate parameters of such an ODE model from the data. However, note that both the structure and order of an ODE model can in principle be learned from data as shown in \cite{ode_ts,em_dim,am,bm}. The remainder of this paper is organised as follows: In Section \ref{methodology}, we present the proposed approach, including the way used for estimating ODE parameters, generating data using an ODE model, and LSTM-AD based anomaly detection. Section \ref{exprmnts_and_results} presents experiments and results. Lastly, we provide conclusion in Section \ref{conclude}.

\section{Proposed Approach}
\label{methodology}
Consider a multivariate time-series $X=\{\mathbf{x}^{(1)},\mathbf{x}^{(2)},...,\mathbf{x}^{(n)}\}$, where each point $\mathbf{x}^{(t)} \in R^m$ in the time series is an $m$-dimensional vector of variables/sensors. Time series data arising from a controlled dynamical system can usually be divided into two sets of variables: $X_{c}$ being a set `control' variables such as accelerator pedal position (APP), and $X_{d}$ which represents `dependent' variables such as coolant temperature (CT), torque, etc. In our application, we observe that the behaviour of control variables was often relatively simple (as depicted in the top panel of Fig.\ref{fig:ode_reconsructed}), which allows us to model them using statistical distributions and to generate synthetic data by sampling these distributions. We learn an ODE model for the dependent variables $X_{d}$ (expressed as a function of $X_c$ and $X_{d}$), using which we generate data for $X_{d}$ corresponding to synthetically produced novel manual control variables $X_{c}$, as explained latter in Section \ref{generate_ODE}. Finally, we make use of ODE-generated time-series data comprising of both sets ($X_{c}$ and $X_{d}$) of sensors to augment the training set for an LSTM-AD model, which can be used to detect anomalous, i.e., non-normal behaviour.
 
\subsection{Learning ODE parameters}
\label{learn_ODE}
An ODE model for a dependent variable $X_{d}$ can be represented as in Eq. \ref{eq:ODE_model}, where $P_{w(t)}$ corresponds to the parameters ($P_{w(t)} = P^{0}_{w(t)},P_{w(t)}^{1},P_{w(t)}^{2}$) of the ODE within a time window $w(t)$. The task of learning an ODE model can be divided into two sub-tasks: 1) estimate the structure $f(P_{w(t)},X_{d}(t),X_{c}(t))$, i.e., nature of interaction among dependent and control variables, and 2) estimate the parameters $P_{w(t)}$. For first sub-task, we anticipate that the ODE model is available from domain knowledge. For example, an ODE structure to model coolant temperature (CT) is mentioned in Eq. \ref{eq:ODE_model} (where $X_{d}=CT$ and $X_{c}=APP $), and similar models are also used in \cite{LFM}. The second sub-task, i.e., estimating $P_{w(t)}$ is non-trivial as the gradients of time-series of sensors present in $X_{d}$ are unknown (LHS of Eq. \ref{eq:ODE_model}). We utilize numerical gradient approximation by Taylor-series expansion to estimate LHS of Eq. \ref{eq:ODE_model}, which reduces the task to simple regression which can be handled using stochastic gradient descent \cite{fast_sgd}, provided $df/dP$ can be computed. However, numerical gradient approximation is not very robust at points with high curvature and/or noise, and a small error in parameter estimation can cause solutions to diverge while integrating ODE models. Also, we have a relatively small number of parameters to estimate when compared with the number of available data-points. Therefore, we smooth the data and drop some of the data-points from the high curvature (sum of first few numerically computed derivatives) regions of time-series.

We optimize the parameters to minimize the root mean square error between the original data and the data obtained after numerical integration of the ODE model. One can further refine the parameters using a gradient-free optimization like particle swarm optimization (PSO). We suggest to initialise PSO using multiple solutions obtained by gradient learning. These solutions correspond to data-points obtained by dropping high curvature points from the time-series data. These solutions can also help in restricting the search space, and therefore, reduce the search time taken by PSO. (Note that for the example in Eq. \ref{eq:ODE_model} and the data used in this paper, we found that the refinement by PSO on initialized solutions was not significant, as the ODE structure is relatively simple.)

\begin{equation}
\label{eq:ODE_model}	
\frac{dX_{d}(t)}{dt} = f(P_{w(t)},X_{d}(t),X_{c}(t))\, ;\,\,\, e.g., \,\,\, \frac{dCT(t)}{dt} = P^{0}_{w(t)}*APP(t) - P^{1}_{w(t)}*CT(t) + P^{2}_{w(t)}  
\end{equation}

\subsection{Using ODE to Generate Data}
\label{generate_ODE}

\begin{figure}
\centering
\includegraphics[width=0.75\textwidth,height=3.5cm]{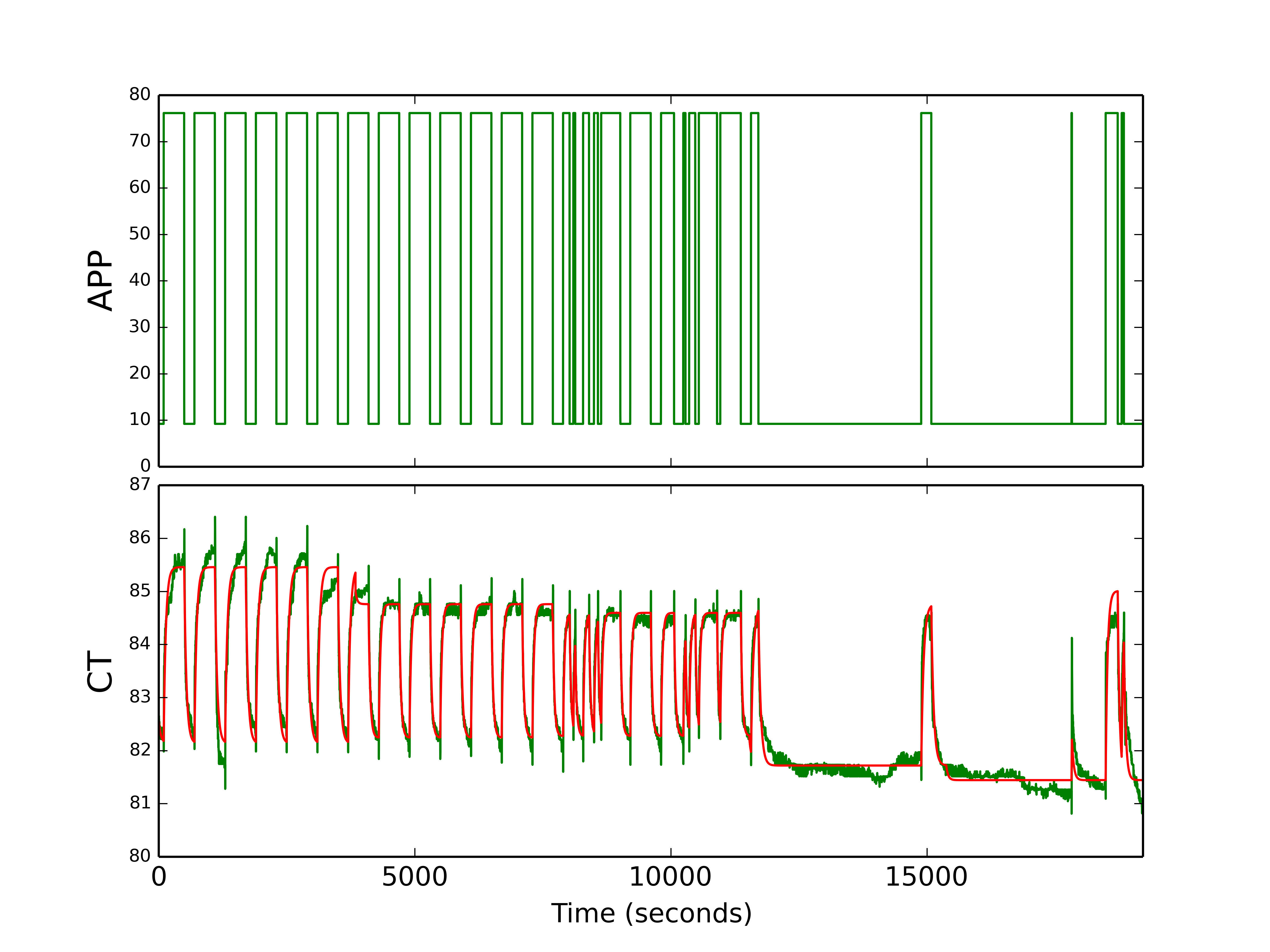}
\caption{Sample data for accelerator pedal position (APP), ODE-generated (red) and real (green) coolant temperature (CT) arising from engine application. Figure best viewed magnified.}
\label{fig:ode_reconsructed}
\end{figure}

As mentioned in Section \ref{learn_ODE}, the behaviour of the control variable(s) ($X_c=\{APP\}$) is often relatively simple, e.g., the APP operates in two states: `high' and `low' as depicted in the top panel of Fig. \ref{fig:ode_reconsructed}. We learn statistical distributions, i.e., histograms for both duration and level of each state using available training data and further sample these distributions to generate novel manual control inputs. For dependent variable(s) ($X_d=\{CT\}$), we numerically integrate the learned ODE model for novel control inputs(s) ($X_c=\{APP\}$). The ODE parameters are learnt for all the time-series pairs ($APP$,$CT$) available in the training and one such pair is used in generating the data for a novel control input(s) ($X_c=\{APP\}$). We have used Euclidean distance (between durations and level of both states, after normalization as they have different scales) as a criterion to select one such pair. 

\subsection{LSTM-AD: LSTM-based Anomaly Detection}
\label{LSTM_AD}
We learn a stacked / deep LSTM based prediction model where $\mathbf{x}^{(t)}$ is used to predict \{${\mathbf{x}^{(t+1)},...,\mathbf{x}^{(t+l)}}$\}, i.e., time-series for next $l$ time-steps. An \textit{error vector} $\mathbf{e}^{(t)}$ for point $\mathbf{x}^{(t)}$ is given by $\mathbf{e}^{(t)}=[e^{(t)}_{1},e^{(t)}_{2},...,e^{(t)}_{l}]$, where $e^{(t)}_{i}$ is the difference between $\mathbf{x}^{(t)}$ and its value as predicted at time $t-i$. The likelihood of a point $\mathbf{x}^{(t)}$ being normal is given by the likelihood score of the corresponding error vector $\mathbf{e}^{(t)}$ computed from a learned Gaussian probability density function over the set of error vectors from the normal data. The parameters of the Gaussian distribution are estimated using Maximum Likelihood Estimation over a set of error vectors from normal time-series data. Further, a threshold on likelihood scores is estimated by maximizing F-score so that points with likelihood score below the threshold are considered to be anomalous points. Different validation sets were used to avoid over-fitting while learning network parameters, prediction length and threshold for likelihood scores. For further details on LSTM-AD, readers can look at \cite{LSTM_AD}.

\section{Experiments and Results}
\label{exprmnts_and_results}

\begin{figure}
\centering
\includegraphics[width=0.75\textwidth,height=5cm]{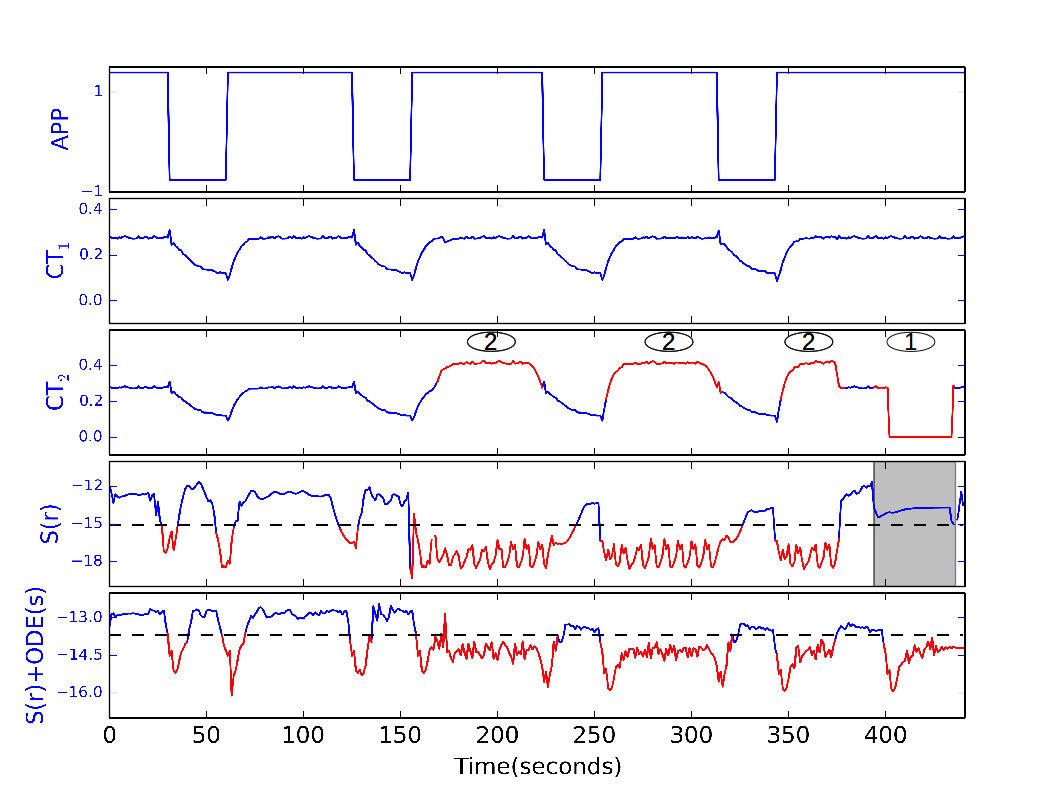}
\caption{From the top, panel-1, panel-2 and panel-3 contains time-series for APP, $CT_{1}$, i.e., normal CT prior to anomaly insertion , and $CT_{2}$, i.e., anomalous CT obtained after injecting anomalies (type 1 \& 2)  respectively. Panel-4 and panel-5 show log-likelihoods (and thresholds) of model learned using small real (S(r)) dataset and its ODE-augmented dataset (S(r)+ODE(s)).}
\label{fig:likelihood-score}
\end{figure}

\begin{figure}
\begin{floatrow}
\ffigbox{%
	\resizebox{60mm}{18mm}	
	{\includegraphics[scale=1]{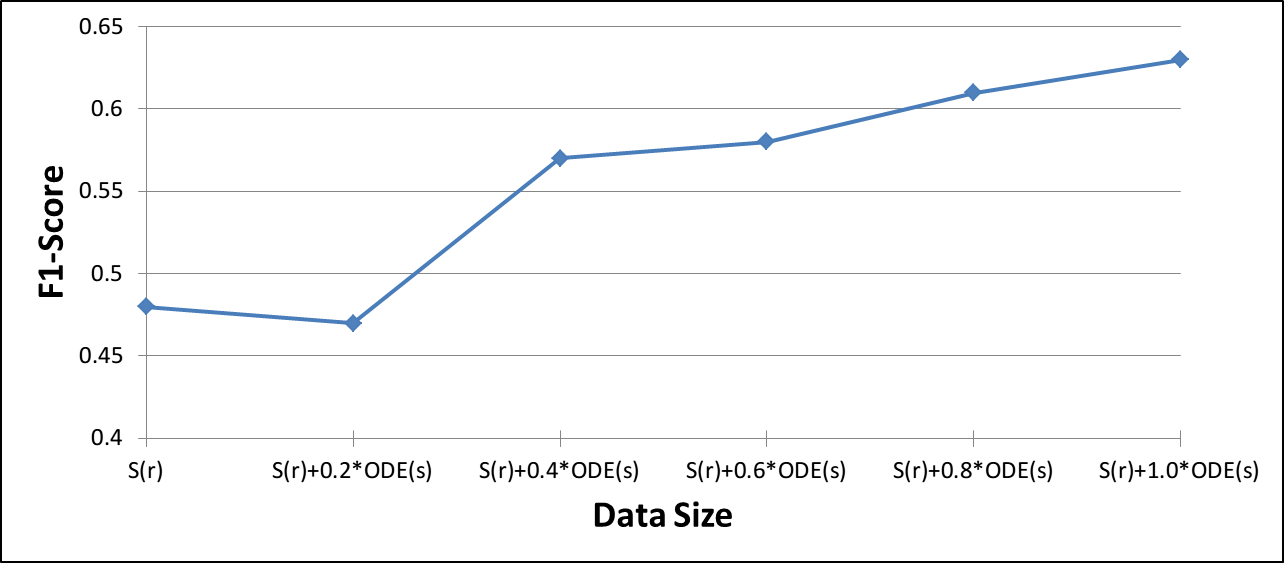}}
}{%
	\label{fig:F-score}

  	\caption{F-Score with respect to addition of ODE-generated data (ODE(s)) to real small dataset (S(r)). Figure best viewed magnified.}%
}
\capbtabbox{%
\scriptsize{
\begin{tabular}{|l|l|l|l|l|l|}
\hline
Train datasets & NS &   NP      & P & R & F \\ \hline
L(r)       & 322 & 34823 & 0.41      & 0.84   & 0.55    \\ \hline
S(r)       & 40 & 3571 & 0.34      & 0.85   & 0.49    \\ \hline
ODE(s)    & 125 & 13598 & 0.32      & 0.82   & 0.46    \\ \hline
S(r)+ODE(s) & 165 &17169 & 0.52      & 0.83   & 0.64    \\ \hline
L(r)+ODE(s) &447& 48421 & 0.56      & 0.84   & 0.67    \\ \hline
\end{tabular}
}
}{%
	\label{tab:ode_prf}
	\scriptsize{
  	\caption{Precision(P), Recall(R) and F-scores(F) for original real Large/Small data (L(r)/S(r)) and ODE-generated data (ODE(s)). NS/NP implies no. of time-series/data-points. }
  	}
}
\end{floatrow}
\caption{testing}
\end{figure}

We evaluate performance of the proposed approach on a sensor data arising from a real world vehicle-engine application. Two sensors in this dataset are APP and CT, where the former is a manual control sensor and the latter is a dependent sensor. We inject different types of anomalies in the time-series recorded for normal CT operation to obtain data for anomalous behaviour  \cite{anomalyDetection_multi_dim_ts}. The following anomaly types were inserted: 1) sudden occurrence of zero value for short duration, 2) value being higher/lower than the normally observed range, 3) dependent sensor CT deviating from expected behaviour, i.e, CT behaving as if control sensor is low when it was actually high, 4) noise in a randomly selected part of the time-series, and 5) gradual increase in values of CT sensor beyond the maximum observed value during normal behaviour. Instances of type 1 and type 2 are shown in (red coloured portion in third panel of) Fig. \ref{fig:likelihood-score}. We further introduce semantic information by injecting anomalies in the randomly selected regions of high state of control sensor, except for anomaly type 4, which was injected in the randomly selected regions of both the states. To show the value of data generated using the ODE model, we train LSTM-AD model with five different training datasets, namely, real large/small data (L(r)/S(r)), ODE-generated (ODE(s)), ODE-augmented large/small data (L(r)/S(r) + ODE(s)). For every training dataset, we tried several deep and shallow architectures of LSTM and chose the one with best performance on the validation set. Results in Table-$1$ show improvements in precision and F-score using ODE-augmented datasets(S(r)/L(r) + ODE(s)) when compared to the real data (S(r)/L(r)). Also in Fig.\ref{fig:F-score}, F-score has improved with progression of the ODE-generated data into to S(r). Fig.\ref{fig:likelihood-score} shows an instance of type 1 anomaly missed (detected) when the model is trained on S(r) ( S(r)+ODE(s) ), as highlighted the grayed out region.

\section{Conclusion}
\label{conclude}
Insufficient data can pose challenges for training machine learning algorithms for anomaly detection, particularly for deep models. In this paper, we leverage ODE models from domain knowledge to augment time-series data arising from dynamical systems. The augmented time-series data was shown to be useful in learning normal behaviour more robustly. We claim the benefits of the ODE-augmented data is due to the fact that it contains sufficient data for the normal behaviour of a dynamical system with a larger collection of variations in the manual control input(s). In future, it will be interesting to investigate the possibility of a purely data-driven ODE based approach \cite{ode_ts,em_dim}.

\bibliographystyle{unsrt}
\bibliography{esann}
\end{document}